\pdfoutput=1
\documentclass[11pt]{article}
\usepackage[final]{coling}

\usepackage{times}
\usepackage{latexsym}

\usepackage{amssymb}
\usepackage[T1]{fontenc}
\usepackage[utf8]{inputenc}

\usepackage{microtype}

\usepackage{inconsolata}

\usepackage{graphicx}

\usepackage{booktabs}
\usepackage{multirow, multicol}
\usepackage{amsmath}

\usepackage{graphicx}
\usepackage{subcaption}

\title{Leveraging Retrieval Augment Approach for Multimodal Emotion Recognition Under Missing Modalities}

\author{
 \textbf{Qi Fan \textsuperscript{1}} \thanks{Equal contributions.},
 \textbf{Hongyu Yuan \textsuperscript{1} \thanks{Equal contributions.}},
 \textbf{Haolin Zuo \textsuperscript{1}},
 \textbf{Rui Liu\textsuperscript{1}} \thanks{Corrsponding author.},
 \textbf{Guanglai Gao\textsuperscript{1}},
\\
\\
 \textsuperscript{1}Inner Mongolia University,
\\
 \small{
   \textbf{Correspondence:} \href{liurui\_imu@163.com}{liurui\_imu@163.com}
 }
}

\begin{document}
\maketitle
\begin{abstract}
Multimodal emotion recognition utilizes complete multimodal information and robust multimodal joint representation to gain high performance. However, the ideal condition of full modality integrity is often not applicable in reality and there always appears the situation that some modalities are missing. For example, video, audio, or text data is missing due to sensor failure or network bandwidth problems, which presents a great challenge to MER research. Traditional methods extract useful information from the complete modalities and reconstruct the missing modalities to learn robust multimodal joint representation. These methods have laid a solid foundation for research in this field, and to a certain extent, alleviated the difficulty of multimodal emotion recognition under missing modalities. However, relying solely on internal reconstruction and multimodal joint learning has its limitations, especially when the missing information is critical for emotion recognition. To address this challenge, we propose a novel framework of \textbf{R}etrieval \textbf{A}ugment for Missing Modality \textbf{M}ultimodal \textbf{E}motion \textbf{R}ecognition (RAMER), which introduces similar multimodal emotion data to enhance the performance of emotion recognition under missing modalities. By leveraging databases, that contain related multimodal emotion data, we can retrieve similar multimodal emotion information to fill in the gaps left by missing modalities. Various experimental results demonstrate that our framework is superior to existing state-of-the-art approaches in missing modality MER tasks. Our whole project is publicly available on GitHub.
\end{abstract}

\section{Introduction}
Multimodal Emotion Recognition (MER) aims to recognize and understand human emotions by integrating multiple modalities such as text, audio, and video. By fusing information from different modalities, MER provides a more comprehensive and accurate sentiment analysis, which plays a crucial role in applications like video conferencing, virtual assistants, and social media analysis, thereby enhancing the emotional understanding and human-computer interaction experience of systems \citep{moin2023emotion, chandrasekaran2021multimodal}.

However, in real-world scenarios, MER tasks frequently encounter situations where certain modalities are missing due to factors such as sensor failures or network bandwidth issues causing data corruption or loss. This significantly increases the difficulty of accurately understanding emotions \citep{hazarika2020, zhao2021missing}. Existing approaches to address this problem primarily focus on two strategies: (1) reconstructing the missing modality representations, and (2) learning robust multimodal joint representations using the available modalities. For instance, \citet{zuo2023exploiting} employed autoencoders to reconstruct missing modality representations and learn robust joint representations based on modality-specific and modality-invariant features; \citet{liu2024contrastive} utilized contrastive methods to extract modality-invariant features and reconstruct missing modalities under different missing conditions.

Despite these efforts, the negative effects of missing modalities can still adversely affect the performance of multimodal learning, leading to incorrect predictions. This is because most existing methods either assume that the missing modalities can be accurately reconstructed from the available data or that the available modalities contain sufficient complementary information to compensate for the missing ones, effectively eliminating the noise introduced by missing data. However, these assumptions may not always hold, especially when the missing information is substantial or when the available modalities lack sufficient correlation with the missing ones. Consequently, the information loss from missing modalities continues to pose a significant challenge.

To address these limitations, we propose a novel framework that introduces retrieval augmentation into MER under missing modality conditions. By constructing a multimodal emotion feature database and employing a retrieval approach, our method acquires and supplements the lost information by retrieving relevant emotional features from similar instances in the database. This retrieval augmentation does not rely on the possibly flawed assumptions. Instead, it introduces external information that can more effectively compensate for the missing modalities, thereby enhancing the robustness and accuracy of emotion recognition even in the presence of missing or noisy data.

The main contribution of this paper can be summarized as follows:

$\bullet$ We explore a new application area of retrieval augment approach: construct multimodal emotion feature databases, and apply them to enhance multimodal emotion recognition.

$\bullet$ We propose a novel retrieval augment framework for multimodal emotion recognition under missing modalities called ``RAMER'', to introduce the emotion feature retrieval approach to bring more emotion information and reduce the information loss by the missing modalities.

$\bullet$ Experimental results under various missing conditions show that our framework outperforms all the baselines and demonstrates robustness in the face of missing data.

\section{Related Work}

\subsection{MER Under Missing Modalities}
Existing methods for multimodal emotion recognition under missing modalities can mainly summarized into two kinds: generation or reconstruction of missing modalities; and learning robust multimodal joint representations. Notably, these approaches have been used together in recent studies.

\subsubsection{Generation and reconstruction methods} 
This method concentrates on obtaining the feature of missing modalities through available modalities.
\citet{tran2017missing} proposed the Cascaded Residual Autoencoder (CRA) that leverages a residual mechanism within an autoencoder framework to effectively restore incomplete data from corrupted inputs. \citet{Cai2018} utilized an encoder-decoder structure, to generate missing modalities (positron emission tomography, PET) from existing modalities (magnetic resonance imaging, MRI). 
\citet{zhao2021missing} combined the CRA with cycle consistency loss for cross-modal imputation and imagined the features of missing modalities. 
\citet{fan2023learning} used a variational autoencoder (VAE) to reduce the effect of noise and generate multimodal joint representations.
\citet{liu2024contrastive} proposed a contrastive learning method to acquire modality-invariant features and used for missing modality imagination.

\subsubsection{Multimodal joint representation learning} 
This approach leverages robust multimodal joint representation to mitigate the negative effects of missing modalities
\citet{pham2019found} took incomplete utterances as input and learned utterance-level representations through cyclic translation to ensure robustness to the missing modalities.
\citet{zuo2023exploiting} learned robust multimodal joint representation with modality-specific and modality-invariant features. 
Graph network was used to capture the speaker and time dependence, which improves the accuracy of conversation emotion recognition under the condition of missing modalities situations \cite{lian2023GCNet}.

\subsection{Retrieval augment methods}

Retrieval-Augmented (RA) methods have received increasing attention in recent years, effectively addressing some of the limitations of generative models, such as knowledge updating and long-tailed data coverage, by introducing a retrieval mechanism and integrating external knowledge or context into the generation process \cite{rag-benchmark, rag-nlp}. Initially, RA methods were mainly applied to text generation tasks like question answering, dialogue systems, and summarization \cite{rag-survey}. By incorporating retrieved relevant documents during the generation phase, these methods produce more accurate and informative content, especially in scenarios requiring large amounts of external knowledge.
Beyond the textual domain, RA methods have also made significant progress in multimodal tasks involving audio and video. In audio tasks, the quality of audio generation is enhanced by retrieving relevant acoustic features or textual descriptions \cite{rag-audio}. In video tasks, the quality of subtitle generation for first-person videos can be improved by retrieving relevant third-person videos \cite{rag-video}.

Previous research has shown that the retrieval augment approach can be effectively applied to multimodal inputs by identifying and leveraging relevant content for each modality. In this paper, we introduce an innovative approach that involves constructing an emotional feature database and applying retrieval-augmented methods to query emotional features. This approach can address information loss due to missing modalities, providing a novel solution for enhancing multimodal tasks.

\begin{figure*}[htbp]
\centering
    \includegraphics[width=1.0\linewidth]{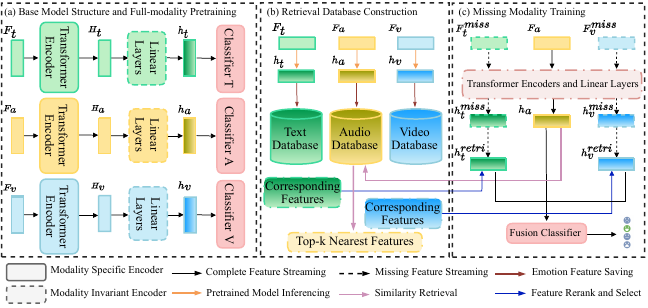}
    \caption{The main structure of our RAMER framework, which includes three stages. For ``Corresponding features'', we acquire Audio top-K similar features and then index the corresponding features with the same sample name from the other two databases.}
    \label{fig:enter-label}
\end{figure*}

\section{Methodology}

\subsection{Data Preparation}

\subsubsection{Dataset}
We utilize the MER2024 multimodal emotion dataset \cite{lian2024mer} as the data source for our research, which is an extension of the MER2023 dataset \cite{lian2023mer}.

% \textbf{IEMOCAP} is a multimodal dataset widely used in emotion recognition research, including speech, video, and text.   The dataset consists of performances of 10 actors in situations and scripted dialogue, covering a variety of emotional categories such as anger, happiness, and sadness. We follow the processing in previous work \cite{zhao2021missing} to process four-class emotion recognition.

\begin{table}[]
\centering 
% \small
\begin{tabular}{lcc}
\toprule
Emotion    & Labeled & Unlabeled \\ \midrule
Happy       & 1,038    & Unknown\\
Angry        & 1,208    & Unknown\\
Sad          & 730     & Unknown\\
Neutral     & 1,248    & Unknown\\
Worried        & 616     & Unknown\\
Surprise       & 190     & Unknown\\ \midrule
Total        & 5,030   & 115,595\\ \bottomrule
\end{tabular}
\caption{The category distribution of the MER2024 dataset. ``Unknown'' indicates this part of data in the dataset has not been labeled and counted by the owner.}% ``-'' means the dataset does not contain this category of emotion.}
\label{category distribution}
\end{table}

MER2024 is an extensive Chinese multimodal emotion dataset comprising 6 emotion categories, 5,030 labeled samples, and 115,595 unlabeled data samples. Table \ref{category distribution} shows the specific emotion distribution of the dataset. MER2024 dataset includes a wide range of film and TV show clips, which have been carefully selected to enhance the diversity and representativeness of the data.  Additionally, filtering steps are employed to reduce noise and eliminate irrelevant content, thereby improving data quality and the accuracy of emotional labeling.  The emotion labels are assigned by multiple emotion recognition experts through a rigorous labeling and cross-verification process, ensuring the high quality and reliability of the dataset.

We employ pretrained language/vision/audio models to obtain text/video/audio embeddings. 
The preprocessing of the dataset and feature extracting process are placed in the Appendix \ref{appendix}.
Finally, we obtain the embeddings with audio, visual, and text modality, noted as $\mathbf{F}_a$, $\mathbf{F}_v$, and $\mathbf{F}_t$.

\begin{table}[]
\centering
\begin{tabular}{l@{\hspace{6pt}}c@{\hspace{6pt}}c}
\toprule
Part Name    & Data Scale & Data Type                       \\ \midrule
% IEMOCAP     & 5531       & labeled data         \\
MER\_small  & 5,030       & labeled data         \\
MER\_medium & 57,780      & unlabeled data       \\
MER\_large  & 115,595     & unlabeled data       \\
MER\_turbo  & 120,625     & all data in MER2024\\ \bottomrule
\end{tabular}
\caption{Various scales of dataset separation.}
\label{various scales}
\end{table}

\subsubsection{Dataset division}

\begin{figure}
    \centering
    \includegraphics[width=1.0\linewidth]{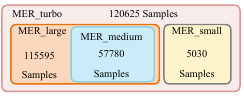}
    \caption{Inclusion relationships between data.}
    \label{fig:scales}
\end{figure}

We divide the MER2024 dataset into four scales, which are listed in Table \ref{various scales}. The inclusion relationships between different parts are exhibited in Fig. \ref{fig:scales}. ``MER\_small'' contains all the labeled data (5,030 samples) in the MER2024 dataset. This part of the data can identify distinct emotional categories and is reliable \citep{lian2023mer}. ``MER\_medium'' consists of about half of all the unlabeled samples (57,780/115,595), which are randomly selected. This scale balanced the size of the retrieval database and the retrieval time. 
``MER\_large'' includes all the unlabeled samples (115,595) in the MER2024 dataset.  ``MER\_turbo'' involves all the labeled and unlabeled data (120,625 samples).

\subsection{Retrieval Augment MER Under Missing Modalities}
Our approach is based on the assumption that, in most cases, different modalities in multimodal emotion data express correlated and similar emotions rather than irrelevant ones. Under this circumstance, we employ the retrieval approach to introduce more emotional information to help emotion recognition and reduce the negative influence of information loss.
The whole structure of our \textbf{RAMER} framework is illustrated in Fig. \ref{fig:enter-label}, which includes three stages: Full-modality pretraining, retrieval vector store construction, and missing modality training.

In the full-modality pretraining stage, we employ the complete labeled data to train a base MER model, ensuring that the model captures comprehensive unimodal emotion features. 
In the second stage, we use the pretrained model saved in the first stage to reason about the entire dataset (including labeled and unlabeled data) and save the unimodal emotion hidden features before each unimodal classifier.
Finally, we proceed to the missing-modalities training stage, where we train the model to handle scenarios with missing modalities by leveraging the retrieval vector store. This allows the model to effectively predict emotions even when some modalities are unavailable.

\begin{table*}[]
\centering \small
\renewcommand{\arraystretch}{1.5} 
\begin{tabular}{l@{\hspace{5pt}}c@{\hspace{5pt}}c@{\hspace{8pt}}c@{\hspace{5pt}}c@{\hspace{8pt}}c@{\hspace{5pt}}c@{\hspace{8pt}}c@{\hspace{5pt}}c@{\hspace{8pt}}c@{\hspace{5pt}}c@{\hspace{8pt}}c@{\hspace{5pt}}c@{\hspace{8pt}}c@{\hspace{5pt}}c@{\hspace{8pt}}c}\toprule
\multicolumn{1}{l}{\multirow{3}{*}{Systems}} & \multicolumn{14}{c}{Testing conditions} \\ \cline{2-15}
\multicolumn{1}{c}{} & \multicolumn{2}{c@{\hspace{15pt}}}{a} & \multicolumn{2}{c@{\hspace{15pt}}}{v} & \multicolumn{2}{c@{\hspace{15pt}}}{l} & \multicolumn{2}{c@{\hspace{15pt}}}{av} & \multicolumn{2}{c}{al} & \multicolumn{2}{c@{\hspace{15pt}}}{vl} & \multicolumn{2}{c@{\hspace{15pt}}}{Avg} \\
\multicolumn{1}{c}{} & WA & UA & WA & UA & WA & UA & WA & UA & WA & UA & WA & UA & WA & UA \\ \midrule
AE & 71.80  & 67.51 & 58.46 & 47.37 & 55.12 & 48.06 & 75.38 & 67.35 & 73.66 & 74.29 & 72.01 & 65.10  & 67.74 & 61.61 \\
CRA & 73.25 & 68.65 & 61.90 & 50.44 & 56.48 & 48.80  & 76.61 & 68.30  & 75.06 & 75.14 & 72.76 & 66.32 & 69.34 & 62.94  \\
MMIN & 74.88 & 70.34 & 62.78 & 50.98 & 57.45 & 50.25 & 77.42 & 69.24 & 75.48 & 75.77 & 73.28 & 67.17 & 70.07 & 63.88 \\
IF-MMIN & 68.79 & 56.81 & 62.35 & 52.38 & 48.79 & 39.81 & 72.94 & 61.07 & 74.51 & 61.55 & 69.12 & 57.67 & 66.23 & 54.96 \\
CIF-MMIN & 72.31 & 69.86 & 64.12 & 52.19 & 63.29 & 62.45 & 79.64 & \textbf{73.93} & 76.38 & 75.61 & 76.44 & 74.93 & 71.98 & 68.29 \\
\textbf{Ours} & \textbf{85.94} & \textbf{74.88} & \textbf{71.34} & \textbf{59.63} & \textbf{89.84} & \textbf{88.23} & \textbf{82.39} & 70.02 & \textbf{90.96} & \textbf{88.99} & \textbf{90.63} & \textbf{86.77} & \textbf{85.13} & \textbf{78.73} \\ 
$\Delta_{sota}$ & 11.06 & 4.54 & 7.22 & 7.25 & 26.55 & 25.78 & 2.75 & -3.91 & 14.58 & 13.22 & 14.19 & 11.84 & 13.15 & 10.44 \\ 
\bottomrule
\end{tabular}
\caption{Main results of baselines and our model on six modality missing conditions. We report WA(\%) and UA(\%)  results and choose WA as the main metric. Both metrics are positively correlated with the model performance. ``a'' means that only modality ``Audio'' is available. ``Avg'' indicates the average result of six conditions. $\Delta_{sota}$ shows the improvement of RAMER compared to state-of-the-art systems.}
\label{main result}
\end{table*}

\subsubsection{Full-modality pretraining}
We first employ three Modality Encoder Networks to capture unimodal emotion features for each modality under complete modalities. They share the same structure of a one-layer Transformer Encoder to encode the embedding, a linear layer to map features to certain dimensions, and a classifier. This stage is trained on the labeled data. After training, we save the best model weights of the epoch that performed best on the test set.

% here
\subsubsection{Retrieval database construction}
We utilize the pretrained model saved in the first stage to infer the whole dataset and respectively save tri-modal emotion hidden features $\mathbf{h}_s$ to three hidden feature databases $\boldsymbol{D_a}$, $\boldsymbol{D_v}$, and $\boldsymbol{D_t}$.  $\mathbf{h}_s$ comes from the output of the last layer before the classifier. Note that the hidden features are aligned according to the sample names, which means that we can search the tri-modal most similar emotion hidden features with one feature at any modality.
Then we select Facebook AI Similarity Search (FAISS) to construct the feature index for $\mathbf{h}_s$ through inner product similarity, which has demonstrated superior performance in various retrieval tasks, particularly when dealing with high-dimensional data \cite{johnson2019billion}. 

\subsubsection{Missing modality training}
\label{training}
In the final stage, we train the model under various missing modalities. Previous research has proved that the addition of full-modality knowledge helps to improve performance in the absence of modalities. Therefore, we first load the pretrained model saved in stage one to leverage its knowledge and send the missing modality data $\mathbf{F}_{s}^{miss}$ into the model and obtain the multimodal hidden feature under missing modalities $\mathbf{h}_{s}^{miss}$. We take text and video modality as missing modalities and the audio modality as an available modality for an example. As Fig. \ref{fig:enter-label} (c) shows, $\mathbf{F}_a$, $\mathbf{F}_{t}^{miss}$, and $\mathbf{F}_{v}^{miss}$ are sent to the model and obtain $\mathbf{h}_a$, $\mathbf{h}_{t}^{miss}$, and $\mathbf{h}_{v}^{miss}$. Then we use the available modality emotion hidden feature $\mathbf{h}_a$ to retrieve the data in $\mathbf{D}_a$ that we built in the second stage, where $\mathbf{D}_a = \{ \mathbf{h}_a^1, \mathbf{h}_a^2, \dots, \mathbf{h}_a^N \}$, $\mathbf{h}_a^i \in \mathbb{R}^d$, $i = 1, 2, \dots, N$.
The cosine similarity is employed to retrieve the database: 
\[
Similarity(\mathbf{h}_a, \mathbf{h}^{similar}_a) = \cos(\theta) = \frac{\mathbf{h}_a \cdot \mathbf{h}^{similar}_a}{\|\mathbf{h}_a\| \|\mathbf{h}^{similar}_a\|}
\]

Then we acquire the top-K similar audio emotion hidden features with feature $\mathbf{h}_a$ from $\mathbf{D}_a$, where
\[
\{\mathbf{h}_a^{sim_1}, \mathbf{h}_a^{sim_2}, \dots, \mathbf{h}_a^{sim_K} \} = \arg\max_{\mathbf{h}_a^i \in D_a}  \cos(\theta_i),
\] 
\[\quad i_1, i_2, \dots, i_K \in \{1, 2, \dots, N\}.
\]

According to the corresponding sample names, we can subsequently acquire the corresponding top-K substitute missing modality multimodal emotional hidden features $\mathbf{h}_{t}^{retris}$ and $\mathbf{h}_{v}^{retris}$, where $\mathbf{h}_{t}^{retris} = \{\mathbf{h}_{t1}^{retri}, \mathbf{h}_{t2}^{retri}, \dots, \mathbf{h}_{tK}^{retri} \}$, and $\mathbf{h}_{v}^{retris} = \{\mathbf{h}_{v1}^{retri}, \mathbf{h}_{v2}^{retri}, \dots, \mathbf{h}_{vK}^{retri} \}$, $K \in \mathbb{Z}^{+}$. Then we fuse the top-K features since the nearest one feature may not share the same emotion with the $\mathbf{h}_a$ feature. We take text modality as an example.
First, we sum the retrieved top-K feature $\mathbf{h}_{t}^{retris}$: \[
\mathbf{h}_t^{\text{sum}} = \sum_{i=1}^{K} \mathbf{h}_{ti}^{\text{retri}}
\]
Then, L2 normalization is performed on the sum vector $\mathbf{h}_t^{\text{sum}}$ to obtain the fused features:
\[
\mathbf{h}_t^{\text{fused}} = \frac{\mathbf{h}_t^{\text{sum}}}{\|\mathbf{h}_t^{\text{sum}}\|}
\]
, where the L2 norm $\|\mathbf{h}_t^{\text{sum}}\|$is defined as:
\[
\|\mathbf{h}_t^{\text{sum}}\| = \sqrt{\sum_{j=1}^{d} (\mathbf{h}_t^{\text{sum}}[j])^2}
\]

Through the fusion method, we can further obtain missing modality data that on the whole is closer to the available modality emotion. Finally, the completed multimodal emotion hidden features $\mathbf{h}_{s}^{comp}$ will be concatenated and sent to the classifier.

\section{Experiments}

\begin{table*}[] \centering
\small
\renewcommand{\arraystretch}{1.4} 
\begin{tabular}
{l@{\hspace{5pt}}c@{\hspace{5pt}}c@{\hspace{8pt}}c@{\hspace{5pt}}c@{\hspace{8pt}}c@{\hspace{5pt}}c@{\hspace{8pt}}c@{\hspace{5pt}}c@{\hspace{8pt}}c@{\hspace{5pt}}c@{\hspace{8pt}}c@{\hspace{5pt}}c@{\hspace{8pt}}c@{\hspace{5pt}}c@{\hspace{8pt}}c} \toprule
\multirow{2}{*}{Systems} & \multicolumn{2}{c@{\hspace{15pt}}}{a} & \multicolumn{2}{c@{\hspace{15pt}}}{v} & \multicolumn{2}{c@{\hspace{15pt}}}{l} & \multicolumn{2}{c@{\hspace{15pt}}}{av} & \multicolumn{2}{c}{al} & \multicolumn{2}{c@{\hspace{15pt}}}{vl} & \multicolumn{2}{c@{\hspace{15pt}}}{Avg} \\
 & WA & UA & WA & UA & WA & UA & WA & UA & WA & UA & WA & UA & WA & UA \\ \midrule
w/o retrieval & 67.81 & 55.54 & 56.12 & 45.72 & 55.96 & 53.30 & 75.13 & 65.29 & 74.75 & 70.95 & 79.49 & 71.60 & 67.81 & 61.55 \\ 
Unimodal $\mathbf{F}_s$ & 63.51 & 50.23 & 51.03 & 40.39 & 53.26 & 49.58 & 69.80 & 60.02 & 69.33 & 67.12 & 73.20 & 64.76 & 63.36 & 55.35 \\
Multimodal $\mathbf{h}_s$ & 70.36 & 62.71 & 62.85 & 50.44 & 64.86 & 60.53 & 73.25 & 64.33 & 75.58 & 73.41 & 76.12 & 70.60  & 70.50  & 63.67 \\
Euclidean & 71.42 & 65.02 & 58.68 & 52.35 & 66.88 & 63.43 & 70.09 & 62.61 & 78.12 & 75.39 & 74.08 & 66.38 & 69.88 & 64.20  \\
MER\_turbo\_top1 & 79.98 & 73.92 & 63.28 & 54.51 & 83.28 & 80.50 & 83.25 & 72.58 & 90.09 & 86.24 & 87.23 & 84.89 & 81.24 & 75.27 \\ \midrule
MER\_small\_top5 & 81.52 & 69.25 & 74.27 & 57.47 & 86.22 & 85.28 & 85.52 & 72.60 & 92.84 & 93.60 & 85.93 & 85.09 & 84.49 & 79.93 \\
MER\_small\_top10 & 85.94 & 74.88 & 71.34 & 59.63 & 89.84 & 88.23 & 82.39 & 70.02 & 90.96 & 88.99 & 90.63 & 86.77 & 85.13 & 78.73 \\
MER\_small\_top15 & 78.87 & 65.33 & 70.94 & 56.26 & 92.63 & 90.21 & 85.68 & 74.36 & 90.71 & 92.12 & 89.98 & 89.22 & 85.06 & 80.72 \\ \midrule
MER\_medium\_top5 & 82.43 & 69.84 & 70.15 & 55.06 & 85.22 & 82.25 & 82.21 & 70.80  & 91.04 & 89.75 & 90.30  & 85.32 & 83.56 & 75.50 \\
MER\_medium\_top10 & 81.56 & 70.26 & 70.69 & 56.13 & 83.86 & 83.39 & 84.40  & 72.95 & 90.88 & 90.46 & 88.71 & 86.23 & 83.35 & 76.57 \\
MER\_medium\_top15 & 82.72 & 72.14 & 68.48 & 54.37 & 86.30  & 85.94 & 82.90  & 71.16 & 88.24 & 87.27 & 88.43 & 85.81 & 82.85 & 76.12 \\ \midrule
MER\_large\_top5 & 82.70 & 69.35 & 70.42 & 54.07 & 85.67 & 82.67 & 84.72 & 71.15 & 93.32 & 91.13 & 91.42 & 87.63 & 84.79 & 77.67 \\
MER\_large\_top10 & 81.40 & 67.36 & 70.78 & 57.35 & 83.73 & 83.31 & 88.06 & 74.18 & 93.28 & 92.27 & 89.07 & 88.20 & 84.56 & 79.70 \\
MER\_large\_top15 & 81.50 & 73.17 & 64.83 & 54.17 & 87.53 & 87.99 & 85.42 & 72.56 & 89.31 & 88.55 & 89.10 & 88.38 & 83.32 & 78.95 \\ \midrule
MER\_turbo\_top5 & 82.87 & 77.80 & 69.48 & 56.22 & 84.66 & 82.68 & 81.35 & 69.93 & 91.90 & 87.62 & 91.98 & 89.72 & 84.18 & 79.29 \\
MER\_turbo\_top10 & 79.22 & 65.78 & 72.82 & 57.08 & 87.18 & 86.63 & 83.75 & 71.72 & 92.14 & 90.53 & 93.78 & 90.84 & 84.77 & 78.59 \\
MER\_turbo\_top15 & 78.05 & 77.45 & 69.48 & 56.80 & 88.98 & 88.65 & 78.72 & 65.89 & 91.38 & 88.26 & 91.43 & 89.45 & 82.89 & 77.55 \\
\bottomrule
\end{tabular}
\caption{The ablation study results of our framework. Data scales have been listed in Table \ref{various scales}. ``top5'' means we fuse top-5 most similar emotion hidden features in the retrieval results. ``top1'' means we only use the most similar feature of the retrieval results.}
\label{ablation results}
\end{table*}

\subsection{Baselines}
To evaluate the performance of our proposed approach, we select the following state-of-the-art missing modality multimodal emotion recognition methods as the baselines. The implementation details are listed in Appendix \ref{appendix_2}.

\textit{AutoEncoder (AE)} \cite{Bengio2006} has been used widely in solving missing modality problems \cite{wong2014imputing}. It used a self-supervised approach to impute missing data from input incomplete data. Following previous work \cite{lian2023GCNet}, we optimize the reconstruction loss of the autoencoder and the classification loss jointly in our implementation.

\textit{Cascaded Residual Autoencoder (CRA)} \cite{tran2017missing} is a strong baseline that combines a series of residual autoencoders to restore missing data from corrupted inputs.

\textit{MMIN} \cite{zhao2021missing} performs well in missing modality problems, which combines \textit{CRA} with cycle consistency loss to learn latent representations of missing modalities. 

\textit{IF-MMIN} \cite{zuo2023exploiting} is an enhancement of \textit{MMIN}. It utilizes modality-specific and modality-invariant features to extend the performance of the \textit{MMIN}.

\textit{CIF-MMIN} \cite{liu2024contrastive} learns modality-invariant features through the contrastive learning method and imagines the missing modalities.

\subsection{Ablation study}
We design several ablation studies to prove the effectiveness of our framework in Table \ref{ablation results}. For the experiment ``w/o retrieval'', we remove the retrieval process in Fig. \ref{fig: structure}c and concatenate the $\mathbf{h}_a, \mathbf{h}_t^{miss}, \mathbf{h}_v^{miss}$ directly. For the experiment ``Unimodal $\mathbf{F}_s$'', we select raw embedding $\mathbf{F}_a, \mathbf{F}_v, \mathbf{F}_t$ instead of the emotion hidden feature $\mathbf{h}_a, \mathbf{h}_v, \mathbf{h}_t$ as the source to construct the vector database. The experiment ``Multimodal $\mathbf{h}_s$'' takes the multimodal fusion strategy in the pretrain stage instead of the unimodal pretraining for comparison and utilizes emotional hidden features for the database construction. For experiment ``Euclidean'', we employ Euclidean distance to retrieve features instead of cosine similarity. Experiment ``MER\_turbo\_top1'' only takes the top1 similar feature while not employing the top-K fusion method in Sec. \ref{training}. 

For the rest of the experiments in Table \ref{ablation results}, we validate the generalization performance of our method through using various scales and scopes of vector databases, also various top-K value selections. 

% These experiments also exhibit the influence of database scale variation.

\subsection{Evaluation metrics}
For both datasets, we follow \citet{liu2024contrastive} and other previous works to employ weighted accuracy (WA) \citep{WA} and unweighted accuracy (UA) \citep{UA} to evaluate performance. Due to class imbalance in the dataset, we choose WA as the preferred metric.

\begin{figure*}[htbp]
    \centering
    \begin{subfigure}{0.32\textwidth}
        \centering
        \includegraphics[width=\linewidth]{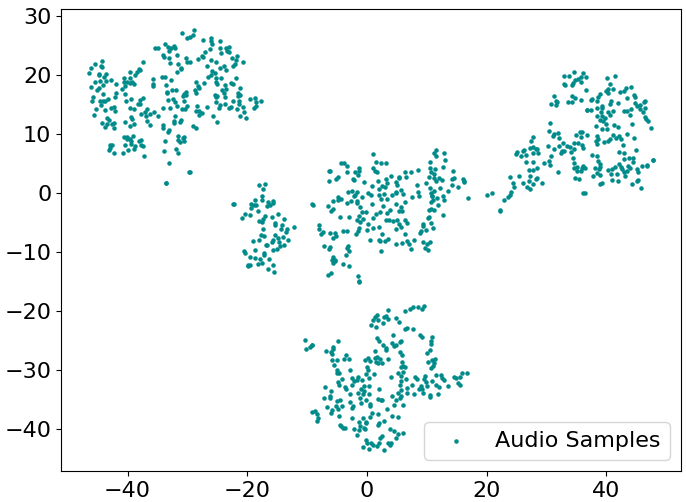}
        % \caption{子图 1 的标题}
        \label{fig:sub1}
    \end{subfigure}
    % \hfill
    \begin{subfigure}{0.32\textwidth}
        \centering
        \includegraphics[width=\linewidth]{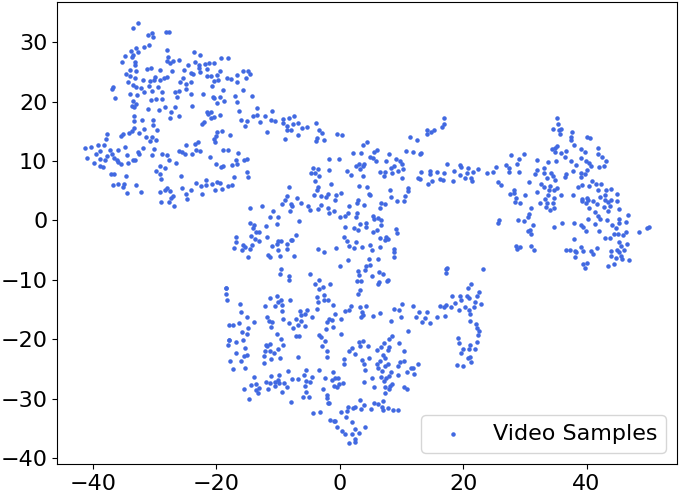}
        % \caption{子图 2 的标题}
        \label{fig:sub2}
    \end{subfigure}
    \begin{subfigure}{0.32\textwidth}
        \centering
        \includegraphics[width=\linewidth]{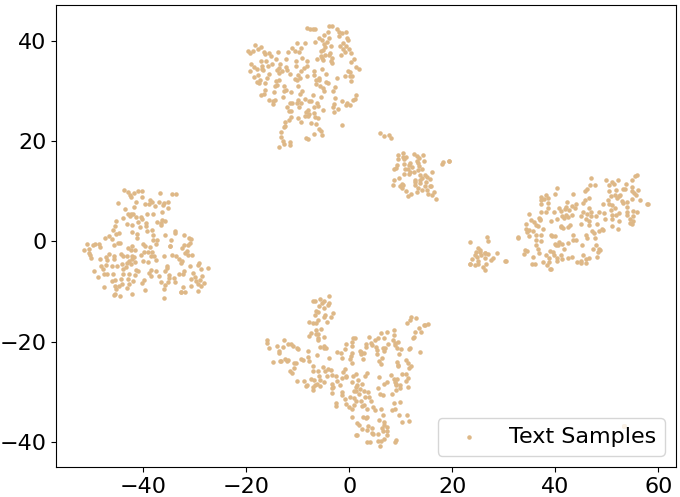}
        % \caption{子图 2 的标题}
        \label{fig:sub2}
    \end{subfigure}
    \caption{The visualization result of 1,000 random samples with each modality in the dataset.}
    \label{fig:main}
\end{figure*}

\section{Results}

\subsection{Main results}
Table 3 shows that our model achieves significant performance improvements under all missing modality conditions.  The performance gains are particularly notable when the text modality is involved (l, al, vl), indicating that the text modality provides rich emotional information and enables the retrieval of more reliable emotional features.  The audio modality follows in importance.  For example, in the last two rows of Table 3, the model achieves the largest improvement, reaching 26.55\%, when only the text modality is available.

Even when only the audio or video modality is available, the model still shows over a 7\% improvement compared to the baselines, demonstrating the robustness of our method.  In terms of absolute performance, the WA exceeds 80\% in all cases except when only the video modality is available and even surpasses 90\% in the al and vl conditions.  This demonstrates that our method exhibits strong performance on most missing modality conditions. 

Based on the results from retrieval databases of different scales, the ``small'' database, despite its limited size, offers higher data quality, leading to the greatest performance (85.13\%).  In contrast, the performance of the ``medium'' database declines compared to the ``large'', indicating that the size of the retrieval database has a certain impact on performance. Both the ``large'' and ``turbo'' databases contain substantial amounts of unlabeled data, which may include instances that negatively affect performance, resulting in lower performance levels. Overall, data quality remains a significant factor influencing retrieval performance, and this quality is largely derived from high-quality manual selections and annotations, making large-scale applications challenging. In practical scenarios, it may be advisable to use ``medium'' or ``large'' databases to balance performance and cost.

\subsection{Ablation results}
From Table \ref{ablation results}, experiment ``w/o retrieval'' in row 3 exhibits the outstanding performance of our approach. On row 4 and 5, the utilizing of unimodal $\mathbf{F}_s$ and multimodal $\mathbf{h}_s$ shows the significant influence of source feature when constructing the retrieval database. The results indicate that directly using raw embeddings to create a retrieval database performs poorly. This may be due to the fact that raw embeddings tend to capture shallow features such as semantics, rather than deeper features like emotions. To effectively capture emotional tendencies, it is necessary to train deep models and apply emotion-related constraints to the learned features. The experiments also reveal that multimodal emotional hidden features do not perform well as unimodal ones in retrieval tasks. This could be attributed to two main reasons: (1) Emotional features are relatively high-level and complex, and the multimodal hidden features learned by current pre-trained models are insufficient for retrieving similar emotions. (2) The modality gap between different types of data makes cross-modal alignment challenging. As a result, the incorporation of multimodal interactions during pre-training may cause emotional features to become confused in unimodal retrieving, leading to a decrease in retrieval accuracy. Row 6 on Table \ref{ablation results} highlights the effect of similarity algorithm selection. Similarity algorithms that are not suitable for data distribution and type may result in a significant decrease in retrieval accuracy.
The experiment in row 7 indicates that the top1 similar feature sometimes may also not share similar emotions with the query feature and lead to misjudgment. 
However, the fusion strategy in Sec. \ref{training} mitigates this situation: one similar feature may deviate from the original emotion, but more similar features may bring the retrieval results back to the correct distribution.

From row 8 to the end, our method demonstrates a fluctuation range of only 2.28\% in WA across different sizes of retrieval databases, varying data scopes, and different top-k values (best performance: 85.13\%, worst performance: 82.85\%). This indicates that our method maintains a considerable level of generalization performance across various retrieval conditions.

\subsection{Visualization}
We randomly select 1,000 samples for each modality in the vector database and visualize their distribution utilizing the T-SNE algorithm. 
From Fig. X we observe that audio and text modality features are approximately clustered into six groups based on emotion categories, with samples of the same category positioned closer together. This demonstrates that our method effectively learns and represents the emotional features of the data.
The clustering effect of the video modality features is less distinct compared to the first two modalities.  This could be because the visual differences between certain emotion categories are not pronounced, for example, surprise and happy, making it more difficult for the model to distinguish between them.

\section{Conclusion}
This paper introduces a novel method for solving multimodal emotion recognition under missing modalities.  We construct multimodal emotional hidden feature databases on the basis of the full-modality pretraining and utilize the retrieval augment approach to fill up the missing modalities and alleviate the information loss. Various experimental results exhibit the advantages of our approach over previous works. We intend to consider learning more robust emotional features and achieve more reliable multimodal emotion recognition under missing modalities.

\section{Limitation}
Despite the promising results of our proposed method, there are two key limitations that need to be addressed in future work. First, the current approach does not take retrieval time into consideration. As the size of the database increases, the retrieval time also grows correspondingly, which may present challenges in practical applications. 
Second, the retrieved content still requires filtering. Current feature fusion strategy may inadvertently incorporate irrelevant emotional features, which can lead to a decline in overall performance. Developing a more accurate feature selection algorithm will be crucial for improving the performance.

\bibliography{custom}

\appendix
\section{Feature extraction}
\label{appendix}
We follow the procedure outlined in the MER2024 Baseline \cite{lian2024mer} and extract utterance-level features. Initially, we preprocess the data by cropping and aligning the facial regions of each frame using the OpenFace toolkit \cite{Baltrusaitis2018}. Next, we employ the CLIP model \cite{radford2021learning} to extract frame-level features for each face image. These visual features are aggregated using average pooling to generate video-level embeddings. For the audio processing, we use the FFmpeg toolkit to separate the audio from the video at a sampling rate of 16 kHz. We then utilize the Chinese-HuBERT-Large model \cite{hsu2021hubert, TencentGameMate} to extract acoustic features, leveraging its superior performance on Chinese sentiment corpora. The final acoustic features are obtained by averaging the hidden representations from the last four layers of the model. Additionally, we transcribe the audio files into text using WeNet \cite{yao2021wenet}, an open-source automatic speech recognition toolkit. For textual feature extraction, we use the Baichuan2-13B-Base model \cite{baichuan2023baichuan2}, which has been pretrained on a large-scale corpus.

\section{Implement Details}
\label{appendix_2}
We split labeled data into train/validation/test sets, and the ratio of data set segmentation is 8:1:1 and they are independent of each other.  We select the best model on the validation set and report the performance on the test set.  For all the experiments, we run five-fold cross-validation three times and take the average result to reduce the impact of random parameter initialization, where each fold contains 40 epochs.  For retrieval, we exclude the query sample from the retrieval results. Additionally, if one retrieval result comes from the validation or test set, it is also excluded to prevent any potential data leakage.

The embedding dimensions for the audio, video, and text features are 1024, 768, and 5120. The hidden size and the saving dimension of the vector store for each modality is 256.  The batch size is set to 128 and the dropout rate is 0.5.

\end{document}